\newcommand{\CLASSINPUTbottomtextmargin}{0.95in}
\documentclass[conference]{IEEEtran}

\usepackage{graphicx}
\usepackage{color}
\usepackage{soul}
\definecolor{light-gray}{gray}{0.85}

\usepackage[linesnumbered,ruled,noend,noline]{algorithm2e} 
\usepackage{cite} 
\usepackage{url}  
\usepackage{ifthen}  
\usepackage{multicol}   
\usepackage{graphicx,courier,amssymb,amsmath,times,caption}
\usepackage{xcolor}
\usepackage{epsfig,psfrag}
\usepackage{bm,graphicx,hhline,amsmath,amssymb,array}
\usepackage{rotating}
\usepackage{amsfonts}
\usepackage{textcomp}
\usepackage{upgreek}
\usepackage{comment}
\usepackage{subfig} 
\usepackage[nolist]{acronym}
\usepackage{bbold}

\usepackage{floatrow}
\acrodef{3GPP}{3rd Generation Partnership Project}
\acrodef{ATG}{air-to-ground}
\acrodef{BS}{base station}
\acrodef{CAM}{Cooperative Awareness Message}
\acrodef{CoMPS}{Continual Meta Policy Search}
\acrodef{CPM}{Collective Perception Message}
\acrodef{DDPG}{Deep Deterministic Policy Gradient}
\acrodef{DQN}{Deep Q-Network}
\acrodef{GD}{gradient descent}
\acrodef{GUE}{ground user equipment}
\acrodef{LoS}{line-of-sight}
\acrodef{MAML}{Model Agnostic Meta Learning}
\acrodef{MDP}{Markov decision process}
\acrodef{NLoS}{non-LoS}
\acrodef{OFDMA}{Orthogonal Frequency Division Multiple Access}
\acrodef{POMDP}{Partially Observable Markov Decision Process}
\acrodef{RB}{Resource Block}
\acrodef{RL}{reinforcement learning}
\acrodef{RSU}{road side unit}
\acrodef{SNR}{Signal-to-Noise Ratio}
\acrodef{SUMO}{Simulation of Urban MObility}
\acrodef{UABS}{unmanned aerial base station}
\acrodef{UAV}{unmanned aerial vehicle}
\acrodef{UMa}{Urban Macro}
\acrodef{V2X}{vehicle-to-everything}
\definecolor{light-gray}{gray}{0.85}

\def\BibTeX{{\rm B\kern-.05em{\sc i\kern-.025em b}\kern-.08em
    T\kern-.1667em\lower.7ex\hbox{E}\kern-.125emX}}

\begin{document}

\title{Continual Meta-Reinforcement Learning for UAV-Aided Vehicular Wireless Networks}

\author{\IEEEauthorblockN{Riccardo Marini\IEEEauthorrefmark{1}, Sangwoo Park\IEEEauthorrefmark{2}, Osvaldo Simeone\IEEEauthorrefmark{2}, Chiara Buratti\IEEEauthorrefmark{1}}

\IEEEauthorblockA{\IEEEauthorrefmark{1}\textit{WiLab, CNIT / DEI, University of Bologna}, Bologna, Italy \\
email: \{r.marini, c.buratti\}@unibo.it}
\IEEEauthorblockA{\IEEEauthorrefmark{2}\textit{KCLIP Lab., CTR, Dept. Engineering, King's College London}, London, United Kingdom\\
email:  \{sangwoo.park, osvaldo.simeone\}@kcl.ac.uk}}

\maketitle

\begin{abstract}
Unmanned aerial base stations (UABSs) can be deployed in vehicular wireless networks to support  applications such as extended sensing via vehicle-to-everything (V2X) services.  A key problem in such systems is designing algorithms that can efficiently optimize the trajectory of the UABS in order to maximize coverage. In existing solutions, such optimization is carried out from scratch for any new traffic configuration, often by means of conventional reinforcement learning (RL). In this paper, we propose the use of continual meta-RL as a means to transfer information from previously experienced traffic configurations to new conditions, with the goal of reducing the time needed to optimize the UABS's policy. Adopting the Continual Meta Policy Search (CoMPS) strategy, we demonstrate significant efficiency gains as compared to conventional RL, as well as to naive transfer learning methods.
\end{abstract}


\section{Introduction}
\label{sec:intro}

Unmanned aerial vehicles acting as flying \acp{BS}, also known as \acp{UABS}, can enhance network capacity by providing on-demand coverage~\cite{3gpp2019uxnb, 3gpp2019uas-services, drones5030094, drones6020039}. 
An important use case is offered by vehicular wireless networks, in which \acp{UABS} serve as relays between vehicular users and the network, enabling the users to upload data collected by on-board sensors \cite{GarMolBobGozColSahKou:21, 9235026, refId0, 8875722, 9700861, 9464915, 9815631}. Such user-generated data are collected by the network, and then forwarded to other vehicles by means of \acp{BS} or \acp{RSU}. Being able to offer stronger, possibly \ac{LoS}, links to vehicles as compared to (static) ground \acp{BS}, \acp{UABS} can support  demanding \ac{V2X} applications, such as advanced driving~\cite{advanceddriving, s20226622} and extended sensing~\cite{DBLP:journals/corr/ChoiPDBH16, s18072207}, as specified by 3GPP~\cite{TS22186}. 
A key problem in such systems is designing algorithms that can efficiently optimize the trajectory of the \ac{UABS} in order to maximize coverage. As a means to find such trajectory, convex optimization approaches have been widely adopted  under the assumption of fixed ground user locations~\cite{jeong2017mobile}. In order to alleviate the impact of the simplifications required to apply convex optimization tools, \ac{RL}-based solutions have been leveraged in~\cite{9659413, 9322234} for the case of static ground users. More challenging scenarios with moving users have been addressed in \cite{cell-free,jiang2021marlhighway, samir2020trajectoryhighway, deng2019jointresourceandtrajectory} using \ac{RL}, where only the speed of the \ac{UABS} was controlled given a fixed trajectory along a highway. The restricted scope of such \ac{RL}-based solutions stems largely from the need to re-train an \ac{RL} policy from scratch for any new environment, e.g., for a new traffic pattern of the ground users.

Therefore, differently from previous works, we propose to mitigate this problem via meta-learning~\cite{thrun1998lifelong}. Meta-learning is able to transfer information from previously experienced configurations to new conditions, reducing the time needed to optimize the \ac{UABS}'s policy. 
Standard meta-learning solutions for \ac{RL}, also known as meta-\ac{RL}, require the designer to have access to the simulators corresponding to all the previously encountered traffic conditions \cite{DBLP:journals/corr/FinnAL17}. This may be practically impossible, or at least computationally prohibitive. Given these limitations of conventional meta-\ac{RL}, this paper explores the use of continual meta-\ac{RL} via \ac{CoMPS}~\cite{DBLP:journals/corr/abs-2112-04467}, which removes the need to revisit previous traffic conditions, and it operates online, acquiring new knowledge as new conditions are encountered. 

Conventional meta-learning was previously considered for \ac{UABS} trajectory optimization in~\cite{hu2020metareinforcement} by assuming that the ground users are static and have known locations. The same authors in~\cite{9457160} extended their previous work by considering multiple \acp{UABS}. 
Unlike these previous works, in this paper, we consider traffic conditions characterized by vehicular users with a priori unknown locations and we move beyond conventional meta-\ac{RL} by accounting for the constraint that simulators for previous traffic configurations cannot be revisited. 
The rest of the paper is organized as follows. The system model and the problem formulation are described in Section~\ref{sec:system_model}. The conventional \ac{RL} framework and the \ac{CoMPS}-based meta-learning scheme are described in Section~\ref{sec:mrl_algo}. Finally, results are presented in Section~\ref{sec:results}, and Section~\ref{sec:conclusion} concludes the paper.
\section{System Model and Problem Formulation}
\label{sec:system_model}
We consider a vehicular network in which an \ac{UABS} provides wireless connectivity to \acp{GUE}. \acp{GUE} produce \ac{V2X} messages that need to be exchanged with the \ac{UABS} in order to provide the network with information related to their surroundings. \begin{figure*}[ht!]
    \centering
    \includegraphics[width=\textwidth]{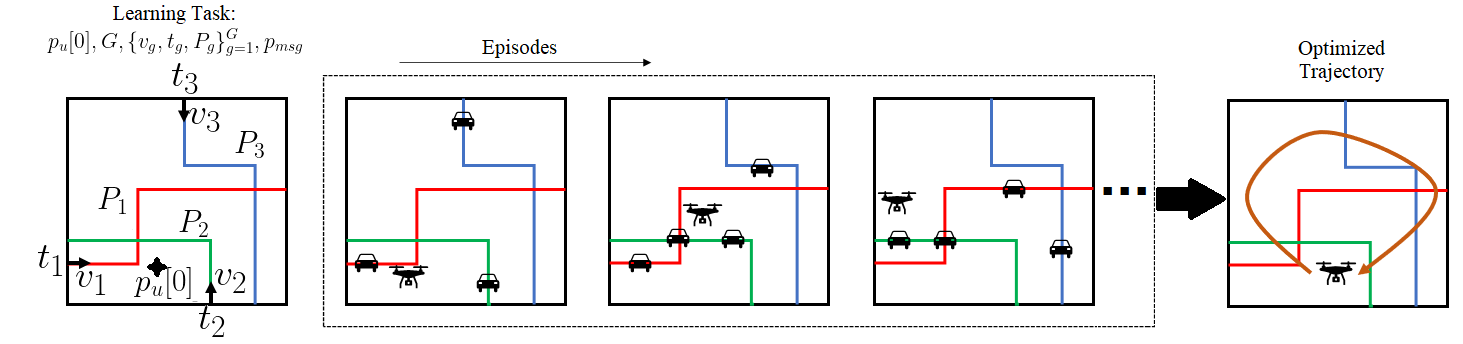}
    \caption{A learning task is defined by an initial \ac{UABS}'s position $p_u[0]$ and by a traffic pattern determined by the number of \acp{GUE}, $G$, the \acp{GUE}' speeds, $\{v_g\}_{g=1}^G$, the \acp{GUE}' discrete starting time instants, $\{t_g\}_{g=1}^G$, paths, $\{P_g\}_{g=1}^G$, and packet generation probability, $p_{msg}$. The \ac{UABS} interacts with the learning task through a simulator over a number of episodes in order to optimize its trajectory.}
    \label{fig:task}
\end{figure*} 
We are interested in optimizing the \ac{UABS}'s trajectory so as to maximize the number of \ac{V2X} packets collected from the \acp{GUE} and relayed to the network during deployment. To this end, we assume access to a simulator configured to mimic current traffic conditions (e.g., generating \acp{GUE}' paths using \ac{SUMO} \cite{SUMO2018}). We aim at reducing the number of episodes that need to be simulated in order to optimize the policy that controls the \ac{UABS}'s trajectory when facing a new task.

\subsection{Learning Task}
\label{sec:learning_task}

As illustrated in Figure \ref{fig:task}, a learning task consists of an initial position $p_u[0]=[x_u[0],y_u[0]]$ of the \ac{UABS} on the plane and of a traffic pattern. 
Time is discretized as $t=0, 1, \ldots, T$, where $T$ is the maximum duration of an episode. 
The traffic pattern is defined by the number $G$ of \acp{GUE}, by the path $P_g$, speed $v_g$ and (discrete) starting time instant $t_{g} \in \{1,\ldots,T\}$ for each \ac{GUE} $g\in\{1,\ldots,G\}$, as well as by the probability $p_{msg}$ that a \ac{GUE} generates a packet at each time step. 
A path $P_g$ is a piece-wise linear curve connecting successive points on the plane.

Given the input parameters $\tau=(G, \{v_g, P_g, t_{g}\}_{g=1}^G, p_{msg})$ defining a traffic pattern, a traffic simulator produces the positions $p_g[t]=[x_g[t], y_g[t]]$ for each \ac{GUE} $g=1, \ldots, G$ at discrete time instants $t=t_{g}, t_{g}+1, \ldots, T_{g}$, where $T_{g}$ is the smaller value between the total duration of an episode, $T$, and the time at which the end point of a path is reached by the \ac{GUE} $g$.  
Specifically, the simulator implements a Markov model $p[t]\sim \mathrm{P}_{\tau}(p[t]|p[t-1])$ to generate the \acp{GUE}' positions $p[t]=[\mathrm{p}_1[t], \ldots, \mathrm{p}_G[t]]$ at time instant $t$  as a function of the previous positions $p[t-1]$ as well as of the traffic pattern $\tau$. The conditional distribution $\mathrm{P}_{\tau}(p[t]|p[t-1])$ can account for interactions among \acp{GUE} and for random events that may affect the \acp{GUE}' trajectories.

Assuming constant altitude, the \ac{UABS}'s position during the $T$ discrete time instants of an episode is described by the sequence $p_u[t]=[x_u[t], y_u[t]]$ for $t \in [0,1,\ldots,T]$. 
At each time instant $t$, the \ac{UABS} can hover, or it can move in one of the eight possible directions $\mathcal{A}_D=\{\leftarrow, \uparrow, \rightarrow, \downarrow, \nwarrow, \nearrow, \searrow, \swarrow\}$. We therefore define the action space $\mathcal{A}=\{\emptyset, \mathcal{A}_D\}$, with $\emptyset$ indicating the hovering decision.

While on route, at each time instant $t \in \{t_g, t_g+1, \ldots, T_g\}$, a \ac{GUE} can produce a message with probability $p_{msg}$. This measurement is stored only for the current time and discarded if not delivered to the \ac{UABS}. Denoting as $\textrm{SNR}_g[t]$ the \ac{SNR} level of \ac{GUE} $g$ towards the \ac{UABS} at time instant $t$, we assume that \ac{GUE} $g$ is \textit{covered} at time $t$ if the inequality
\begin{equation}
\label{eq:snr_th}
    \textrm{SNR}_g[t] \geq \textrm{SNR}_{\mathrm{th}}
\end{equation}
holds, given a fixed threshold $\textrm{SNR}_{th}$. When condition~\eqref{eq:snr_th} is satisfied, the \ac{GUE} can successfully communicate a message to the \ac{UABS} at time instant $t$.
The \ac{UABS} can receive at most $C_{\mathrm{max}}$ packets at the same time $t$. If more than $C_{\mathrm{max}}$ \acp{GUE} satisfy condition \eqref{eq:snr_th} and have a packet to transmit, the \ac{UABS} randomly selects a subset of $C_{\mathrm{max}}$ \acp{GUE} from which to receive a packet.

We aim at optimizing the stochastic policy $\pi(a|s)$ for the \ac{UABS} that selects \emph{action} $a \in \mathcal{A}$ as a function of the current \emph{state}  $s$ of the system, i.e., $a[t] \sim \pi(\cdot|s[t])$. The state is defined as the collection of all positions of \ac{UABS} and \acp{GUE}, $s[t]=(p_u[t], p[t])\in \mathcal{S}$.
After selecting an action $a[t]$, the \ac{UABS} and all the \ac{GUE}s move to state $s[t+1]$ with transition probability $\mathrm{P}_\tau(s[t+1]|a[t],s[t])$ given as
\begin{align}
\label{eq:transition_prob}
    &\mathrm{P}_\tau(s[t+1]|a[t],s[t])\nonumber\\
    &=\mathrm{P}_\tau(p[t+1]|p[t])\cdot \mathbb{1}(p_u[t+1]=f(p[t], a[t])), 
\end{align}
where the conditional distribution $\mathrm{P}_\tau(p[t+1]|p[t])$ is implemented by the traffic simulator; $f(p_u[t], a[t])$ is a function that updates the position of the \ac{UABS} given action $a[t]$; and $\mathbb{1}(\cdot)$ is the indicator function. Given state $s$ and action $a$, the \ac{UABS} obtains a scalar random reward $r[t]\sim \mathrm{P}_{\tau}(r|s)$ equal to the sum of packets collected by the \ac{UABS}, i.e., 
\begin{equation}
\label{eq:reward}
    r=\min\left(C_{\mathrm{max}},\sum_{g=1}^G r_{g} \right).
\end{equation}
In \eqref{eq:reward}, the random variable $r_g$ equals one if \ac{GUE} $g$ has a packet to transmit and satisfies the coverage condition \eqref{eq:snr_th}. 
Note that the random variable $r_g$ is a function of the current state $s$, and that its stochasticity arises from the random packet generation process.

Given an initial \ac{UABS} position $p_u[0]$ and the traffic pattern $\tau$, we formulate the design problem for the policy $\pi(a|s)$ as the optimization of the discounted average return

\begin{equation}
\label{eq:problem_formulation}
\max_{\pi} \left\{ J_{\tau_0}(\pi)= \sum_{t=1}^{T} \gamma^{t}  \mathbb{E}_{\pi(a[t]|s[t])}\left[r[t]\right] \right\},
\end{equation}
with discount factor $\gamma \in (0,1]$~\cite{sutton}. In \eqref{eq:problem_formulation}, we have identified the problem configuration as $\tau_0=[{p}_u[0], \tau]$, and we have made explicit the dependence of the expectation on the policy $\pi(a[t]|s[t])$. 
The average also accounts for the transition probability \eqref{eq:transition_prob} and for the random reward \eqref{eq:reward}.

\subsection{Channel Model}
\label{sec:channel_model}

To define the \ac{SNR} level for each \ac{GUE} $g$, we assume the propagation model described in~\cite{OPTLAP} for an urban environment. Accordingly, links between the \ac{UABS} and \acp{GUE} can either be in \ac{LoS} or \ac{NLoS} conditions. The probability $p_{\textrm{L}g}$ for the link of \ac{GUE} $g$ at time instant $t$ to be in \ac{LoS} condition is
\begin{equation}
p_{\textrm{L}g}[t]=\frac{1}{1+\alpha \exp (-\beta(\theta_g[t] -\alpha))},
\label{eq:plos}
\end{equation} 
where $\alpha$ and $\beta$ are two environment-dependent constants~\cite{OPTLAP}, and $\theta_g[t]$ is the elevation angle for the ray connecting the \ac{GUE} $g$ and the \ac{UABS} at time $t$. 
The path loss between the \ac{GUE} $g$ and the \ac{UABS} at time instant $t$ is given by
\begin{equation}
\mathrm{L}_g[t]=20\log_{10}(f_c) + 20\log_{10}(d_g[t]) - 27.55 +\eta_{\xi,g} \quad[\textrm{dB}],
\label{eq:los}
\end{equation}
with carrier frequency $f_c$ in MHz; distance $d_g[t]$ between the \ac{GUE} $g$ and the \ac{UABS} at time instant $t$ in meters; and excessive path loss coefficient   $\eta_{\xi,g}$ \cite{OPTLAP}, with $\xi$ being a binary index indicating whether the link is in \ac{LoS} or \ac{NLoS} conditions.
Finally, based on \eqref{eq:los}, the \ac{SNR} of \ac{GUE} $g$ at time instant $t$ can be expressed as \cite{OPTLAP}
\begin{equation}
    \mathrm{SNR}_g[t]=(P_{\rm tx} + G_{\rm tx} + G_{\rm rx} - \mathrm{L}_g[t]) - P_{\rm noise} \quad[\textrm{dB}],
    \label{eq:snr}
\end{equation}
where $P_{\rm tx}$ is the transmitted power of \acp{GUE} in dBm; $G_{\rm tx}$ and $G_{\rm rx}$ represent the gain in transmission and reception in dB, respectively; and $P_{\rm noise}$ is the noise power at the \ac{UABS} in dBm.
\section{Meta-Reinforcement Learning Algorithm}
\label{sec:mrl_algo}

In this section, we first introduce the standard reinforcement learning (\ac{RL})-based solution. This approach addresses problem \eqref{eq:problem_formulation} from scratch for a fixed configuration $\tau_0$ given by initial \ac{UABS} position $p_u[0]$ and traffic pattern $\tau$. We then exploit continual meta-learning, capable of transferring knowledge across different configurations, to avoid a large number of training episodes.

\subsection{Conventional Reinforcement Learning}
\label{sec:rl}

To address problem \eqref{eq:problem_formulation} for a given configuration $\tau_0$, we introduce a parameterized policy $\pi_\theta(a|s)$, and we adopt the standard policy gradient method \cite{NIPS1999_464d828b, sutton}. Accordingly, the gradient of the reward function $J_{\tau_0}(\pi_\theta)$ in \eqref{eq:problem_formulation} is estimated as 
\begin{equation}
\label{eq:policy_gradient_estimation}
\widehat{\nabla}_\theta J_{\tau_0}(\pi_\theta)= \sum_{t=0}^{T}\nabla_\theta\log\pi_\theta(a[t]|s[t])G[t],
\end{equation}
with return $G[t]=\sum_{t'=t}^T\gamma^{t'-t}r[t']$. The gradient \eqref{eq:policy_gradient_estimation} is computed at the end of each episode of $T$ time steps based on the \emph{experience} $e:= [s[0],a[0],r[0],\ldots,s[T],a[T],r[T]]$. The gradient \eqref{eq:policy_gradient_estimation} is used to update the policy parameters vector $\theta$ as 
\begin{equation}
\label{eq:policy_update}
\theta \leftarrow \theta + \eta \widehat{\nabla}_\theta J_{\tau_0}(\pi_\theta)
\end{equation}
with learning rate $\eta >0$~\cite{sutton}.

\subsection{Meta-Reinforcement Learning}
\label{sec:ml}

In continual meta-\ac{RL}, the \ac{UABS} explores configurations $\tau_0^i$ sequentially over a discrete index $i=0,1,\ldots$ The goal is to transfer knowledge from previously observed tasks so as to prepare to solve problem \eqref{eq:problem_formulation} for future configurations using fewer episodes. A key challenge in this process is posed by the assumption that the \ac{UABS} cannot run additional simulations for previously encountered configurations. As we will see, this problem can be addressed by storing information about experiences from previous configurations.

\begin{figure}[h!]
    \centering
    \includegraphics[width=0.95\textwidth]{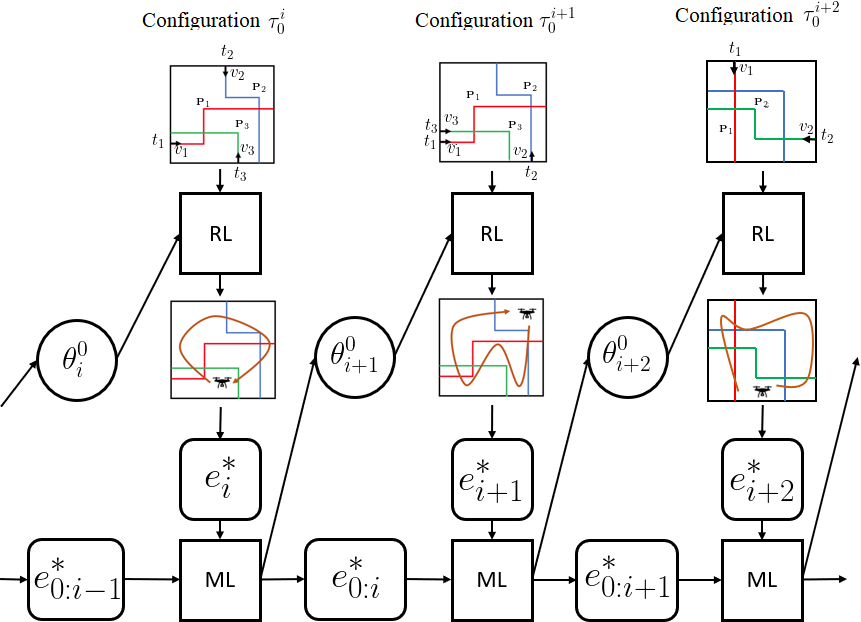}
    \caption{Continual meta-reinforcement learning: For each new configuration $\tau_0^i$ comprising \ac{UABS}'s initial position and traffic pattern, the \ac{UABS} implements \ac{RL} to optimize its trajectory starting from the current initialization of the policy parameter vector $\theta^0_{i}$ inherited from the previous configurations. After completing optimization on the current configuration, experiences are saved in separate sets, and a meta-learning step (ML) is carried out using offline \ac{RL}.}
    \label{fig:mrl_algo}
\end{figure}

Following \cite{DBLP:journals/corr/abs-2112-04467}, we assume that information is transferred from previous tasks in the form of an initialized model parameter vector $\theta^0$ for the policy gradient update \eqref{eq:policy_update}. As illustrated in Figure \ref{fig:mrl_algo}, continual meta-\ac{RL} consists of two main steps applied for each new configuration $\tau_0^i$:

\begin{itemize}
    \item Conventional policy gradient-based \ac{RL} is applied over $N$ episodes to maximize the expected reward $J_i(\theta)=J_{\tau_0^i}(\pi_\theta)$ with initialization $\theta_i^0$, producing the optimized parameter vector $\theta_i^*(\theta_i^0)$ as a function of $\theta_i^0$;
    \item A meta-update of the initialization $\theta_i^0$ is applied with the goal of maximizing the sum of the expected rewards for the configurations encountered so far for the problem
    \begin{align}
        \label{eq:meta-update}
        \theta_{i+1}^0 \leftarrow \arg\max_{\theta^0} \sum_{i'=0}^{i}\Tilde{J}_i (\Tilde{\theta^*_i}(\theta^0)).
    \end{align}
\end{itemize}
In \eqref{eq:meta-update}, the notations $\Tilde{J}_i(\theta)$ and $\Tilde{\theta^*_i}$ indicate that the \ac{UABS} cannot run new episodes for previous and current tasks, and hence it can only estimate the average return ${J_i(\theta)}$ and the optimized model parameter vector $\theta^*_i(\theta^0)$ for configurations $i'=0, \ldots, i$. These are explained next.

In order to estimate $J_i(\theta)$ along with the policy parameter $\theta^*_i(\theta^0)$ without reusing the simulator, for configuration $\tau_0^i$, Continual Meta Policy Search (\ac{CoMPS}) \cite{DBLP:journals/corr/abs-2112-04467} stores a \emph{full experience set} $\mathcal{E}_i= \{ [e_{i,n}, \pi_{i,n}] \}_{n=1}^N$ including all the experiences
\begin{equation}
\begin{split}
\label{eq:exp}
    e_{i,n} &= [s_{i,n}[0], a_{i,n}[0],r_{i,n}[0],\ldots, s_{i,n}[T],a_{i,n}[T],r_{i,n}[T]]
\end{split}
\end{equation}
for configuration $\tau_0^i$, as well as the probabilities to choose the corresponding actions in $e_{i,n}$
\begin{equation}
\begin{split}
\label{eq:policy}
    \pi_{i,n} &= [\pi_{\theta_{i,n}}(a_{i,n}[0]|s_{i,n}[0]), \ldots, \pi_{\theta_{i,n}}(a_{i,n}[T]|s_{i,n}[T])].
\end{split}
\end{equation}
In \eqref{eq:exp} and \eqref{eq:policy}, the notations $s_{i,n}[t], a_{i,n}[t], r_{i,n}[t], \theta_{i,n}$ stand for state, action, reward, and policy parameter at time $t$ for episode $n$ in configuration $\tau_0^i$. 
In addition, the \emph{best} episode $n^*$ is chosen as the episode that achieves the highest total reward without discounting factor $\gamma$ \cite{DBLP:journals/corr/abs-2112-04467}, i.e., $n^*=\arg\max_{n} \sum_{t=0}^{T}r_{i,n}[t]$, and the corresponding experience $e_{i,n^*}$ is saved in the \emph{skilled experience set} $\mathcal{E}_i^*$. 

Using the full experience sets $\{\mathcal{E}_{i'}\}_{i'=1}^{i}$ and the skilled experience sets $\{\mathcal{E}^*_{i'}\}_{i'=1}^{i}$, \ac{CoMPS} addresses problem \eqref{eq:meta-update} as follows. First, \emph{off-policy local updates} are used to obtain the optimized policy parameter vector $\tilde{\theta}^*_{i}(\theta^0)$ as

\begin{align}
\label{eq:off-policy_update}
    \Tilde{\theta^*_{i}}(\theta^0) = \theta^0 &+\eta \sum_{t=0}^{T} \frac{\pi_{\theta^0}(a_{i,n}[t]|s_{i,n}[t])}{\pi_{\theta_{i,n}}(a_{i,n}[t]|s_{i,n}[t])}\nonumber\\
    &\cdot\nabla_{\theta^0}\log\pi_{\theta^0}(a_{i,n}[t]|s_{i,n}[t])G_{i,n}[t]
\end{align}
with learning rate $\eta > 0$ and corresponding discounted return $G_{i,n}[t]=\sum_{t'=t}^T\gamma^{t'-t}R_{i,n}[t']$ as defined in \eqref{eq:policy_gradient_estimation}. In \eqref{eq:off-policy_update}, the episode $n$ is selected at random from the $N$ episodes in set $\mathcal{E}_{i}$. Furthermore, the \emph{importance sampling} ratio $\pi_{\theta^0}(a_{i,n}[t]|s_{i,n}[t])/\pi_{\theta_{i,n}}(a_{i,n}[t]|s_{i,n}[t])$ is included in \eqref{eq:off-policy_update} in order to compensate for the generally different probability assigned to action $a_{i,n}[t]$ given state $s_{i,n}[t]$ by the policies $\pi_{\theta^0}(a|s)$ and $\pi_{\theta_{i,n}}(a|s)$. This can partly mitigate the performance degradation caused by the adoption of off-policy optimization \cite{simeone2022machine, degris2012off}.

The objective $\Tilde{J}_i(\theta)$ is evaluated using the skilled experience $\mathcal{E}_{i}^*$ via behavioral cloning \cite{mendonca2019guided}. The behavioral cloning loss measures how well the policy $\pi_\theta$ can reproduce the near-optimal, skilled trajectory $e_{i,n^*} \in \mathcal{E}_{i}^*$. It is accordingly defined as
\begin{equation}
\label{eq:behavioral_loss}
    \Tilde{J}_i(\theta) = - \sum_{t=0}^{T} \log \pi_\theta(a_{i,n*}[t]|s_{i,n*}[t]).
\end{equation}
Finally, \ac{CoMPS} applies gradient-based optimization to problem \eqref{eq:meta-update} as
\begin{equation}
    \label{eq:meta_update}
    \theta^0 \leftarrow \theta^0 - \frac{\kappa}{i+1} \sum_{i'=0}^{i} \nabla_{\theta^0} \Tilde{J}_i(\tilde{\theta}^*_i(\theta^0)),
\end{equation}
with learning rate $\kappa > 0$. 

In order to reduce computational complexity as $i$ grows in \eqref{eq:meta_update}, we sample $B$ tasks among the available $i+1$ tasks to compute the gradient in \eqref{eq:meta_update}. This way, evaluating the meta-update \eqref{eq:policy_update} requires order  $O(4I_\text{meta}BTC)$ operations, assuming $I_\text{meta}$ iterations for the meta-update \eqref{eq:meta_update}, where $C$ represents the computational complexity of applying policy $\pi_\theta(a|s)$ from the state $s$. In contrast, conventional \ac{RL} \eqref{eq:policy_gradient_estimation} requires order $O(2I_\text{conven}TC)$ operations, where the number of iterations $I_\text{conven}$ is typically very large \cite{9659413}. Therefore, by transferring knowledge from previous environments, meta-RL can significantly reduce the computational complexity.

\section{Experiments}
\label{sec:results}
In this section, we provide insights and experimental evidence on the benefits of meta-learning via \ac{CoMPS} as compared to conventional \ac{RL}. Since meta-learning aims at transferring useful knowledge across different configurations encountered over time index $i$, as a benchmark, we also consider a basic \emph{transfer \ac{RL}} solution, which uses the policy parameter vector $\theta_i^*$ optimized based on the $i$th configuration as the initialization of conventional \ac{RL} (Section \ref{sec:rl}) for the $(i+1)$th configuration. If not stated otherwise, parameters used during the simulations are listed in Table \ref{tab:simulation_parameters}.

\subsection{Toy Example}
We consider first a simple setup consisting of a small 40 m $\times$ 40 m grid world with two possible tasks.
The configurations for the two tasks differ only in the path $P_g$ traveled by the three \acp{GUE} ($G=3$), whereas other parameters are fixed: The initial position of the \ac{UABS} is set as the bottom-right corner of the square area, i.e., $p_u[0]=[20,0]$; the speed for the \acp{GUE} are given as $v_1=v_2=v_3=1$ m per time step $t=1$ s, the message generation probability is $p_{msg}=1$, and the starting time instants of the \acp{GUE} are assumed to be $t_1=1, t_2=2, t_3=3$. The duration of an episode is set to $T=60$ s. In the path $P_g$ for task $\tau_0^1$, all the \acp{GUE} start from the bottom right corner of the square area to move in clockwise direction along the perimeter of the area, while for task $\tau_0^2$ the movement of \acp{GUE} is taken in counterclockwise. Lastly, we assume that the tasks are presented alternatively for every discrete time index $i$. 

\begin{figure}[ht!]
    \centering
    \includegraphics[width=0.95\textwidth]{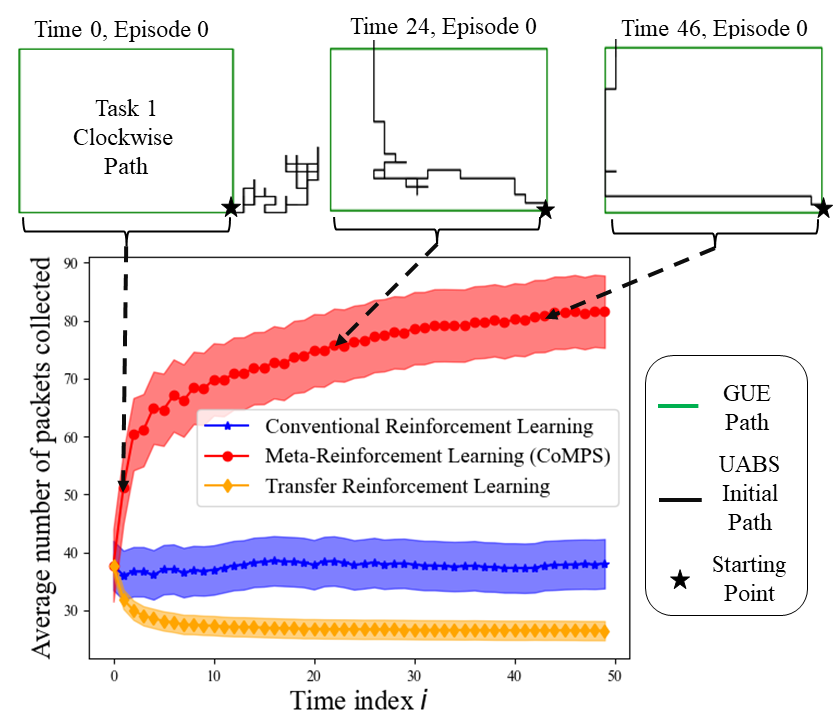}
    \caption{(Bottom) Average number of packets collected by the \ac{UABS} across $N=50$ episodes as a function of time index $i$; (Top) Initial trajectory of \ac{UABS} obtained from the meta-learned initialization $\theta^0_i$ \eqref{eq:meta-update} (visualized as a black line). For this toy example, two tasks are deployed alternately for each time $i$ while the only difference between the two tasks is the path $P_g$: even $i$ takes clockwise path while odd $i$ has counterclockwise path.}
    \label{fig:toy_example_result}
\end{figure}

Fig.~\ref{fig:toy_example_result} plots the average number of packets collected per episode, assuming $N=50$ episodes, over time index $i$. The error regions are obtained by evaluating the standard deviation over 10 independent experiments. Conventional \ac{RL} cannot take advantage of the data from $i$ configurations,  while the performance of transfer \ac{RL} is affected by a negative transfer of information from the previous configurations. In contrast, meta-\ac{RL} via \ac{CoMPS} can effectively transfer information from the $i$ previous configurations. This is illustrated by the initial trajectory optimized by meta-\ac{RL}, which is shown in the top part of Fig.~\ref{fig:toy_example_result} for increasing values of $i$. The figure demonstrates how meta-\ac{RL} gradually identifies a useful initial trajectory from which fast adaptation can be carried out for both tasks.

\subsection{Urban Scenario}


In order to evaluate the effectiveness of meta-learning over a more realistic setting, we simulated traffic patterns using the \ac{SUMO} software for an area in the city of Bologna, Italy, whose dimension is $1500$ m $\times$ $900$ m \cite{SUMO2018}. In this scenario, $K=50$ different task configurations, characterized by different numbers of \acp{GUE} (randomly chosen between 15 and 30) moving with different random speed along different paths, are explored sequentially over time index $i=0,\ldots,49$. The duration of an episode is set to $T=300$ s.

\begin{figure}[h!]
    \centering
    \includegraphics[width=\textwidth]{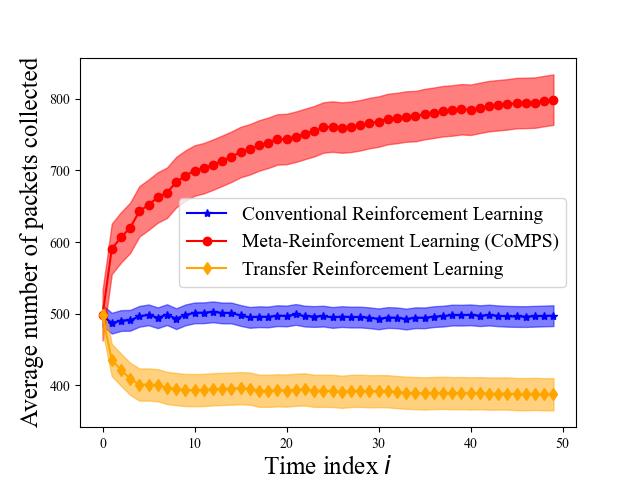}
    \caption{Average number of packets collected by the \ac{UABS} across $N=50$ episodes as a function of time index $i$. \acp{GUE}' paths are generated using the SUMO software \cite{SUMO2018}.} 
    \label{fig:urban_scenario_result}
\end{figure}

Fig.~\ref{fig:urban_scenario_result} shows the average number of packets collected per episode across $N=50$ total episodes as a function of time index $i$. Again, the error regions are obtained by considering the standard deviation over 10 independent experiments. In a manner that reflects well the results reported for the toy example, meta-\ac{RL} outperforms both conventional and transfer \ac{RL} by successfully transferring knowledge from previously encountered configurations.

\begin{table}[h!]
\centering
\caption{Simulation Parameters}
\begin{tabular}{||c|c|c||}
\hline
\textbf{Parameter} & Toy Example & Urban Scenario \\
\hline
$K$ & 50 & 50\\
\hline
$N$ & 50 & 50\\
\hline
$\eta$ & 0.001 & 0.001\\
\hline
$\kappa$ & 0.0001 & 0.0001\\
\hline
$\gamma$ & 0.8 & 0.8\\
\hline
$t$~[s] & 1 & 1\\
\hline
$C_{\mathrm{max}}$ & 10 & 10\\
\hline
$v_{u}$~[m/s] & 1 & 20 \\
\hline
$v_g$~[m/s] & 1 & 10\\
\hline
$P_{\rm tx}$~[dBm] & 0 & 20 \\
\hline
$P_{\rm noise}$~[dBm] & -100 & -100\\
\hline
$G_{\textrm{tx}}$~[dB] & 0 & 0 \\
\hline
$G_{\textrm{rx}}$~[dB] & 0 & 0 \\
\hline
$p_{msg}$ & 1 & 1\\
\hline
$\rm SNR_{th}$~[dB] & 50 & -10\\
\hline
$f_c$~[GHz] & 30 & 30\\
\hline
\end{tabular}
\label{tab:simulation_parameters}
\end{table}
\section{Conclusion}
\label{sec:conclusion}
In this paper, we have addressed the problem of optimizing the trajectory of an \ac{UABS} with the aim of supporting \ac{V2X} services for moving \acp{GUE}. In order to reduce the data requirements for RL-based training, we have proposed to extract useful information from previously encountered traffic configurations to adapt quickly to new environments via meta-RL. Even without the ability to actively revisit previous traffic conditions, we have shown that meta-RL can optimize the initial policy parameter vector so as to reduce the number of exploration steps during training. Future work may consider distributed continual meta-learning across multiple \ac{UABS}.

\section{Acknowledgements} 
The work of R. Marini and C. Buratti was supported by the CNIT National Laboratory WiLab. The work of S. Park and O. Simeone was supported by the European Research Council (ERC) under the European Union’s Horizon 2020 research and innovation programme (grant agreement No. 725731).

\IEEEtriggeratref{0}
\bibliographystyle{IEEEtran}
\bibliography{IEEEabrv, bibliography}

\end{document}